\documentclass[runningheads]{llncs}
\usepackage{graphicx}
\usepackage{tikz}
\usepackage{comment}
\usepackage{amsmath,amssymb} 
\usepackage{color}
\usepackage{hyperref}
\usepackage{svg} 
\usepackage{url}
\usepackage{graphicx}
\usepackage[small]{subfigure}
\usepackage{bbding}
\usepackage{multirow}
\usepackage{xcolor}
\usepackage{booktabs}
\usepackage{makecell}
\usepackage{bbm}
\usepackage{courier}

\newlength\savewidth\newcommand\shline{\noalign{\global\savewidth\arrayrulewidth
		\global\arrayrulewidth 1pt}\hline\noalign{\global\arrayrulewidth\savewidth}}

\definecolor{mydarkblue}{rgb}{0,0.08,0.45}
\hypersetup{
	colorlinks=true,
	urlcolor=magenta,
	citecolor=mydarkblue,
}

\newcommand{\modelname}{{DINO}}

\usepackage[accsupp]{axessibility}  

\usepackage[width=122mm,left=12mm,paperwidth=146mm,height=193mm,top=12mm,paperheight=217mm]{geometry}

\def\@fnsymbol#1{\ensuremath{\ifcase#1\or *\or \dagger\or \ddagger\or
   \mathsection\or \mathparagraph\or \|\or **\or \dagger\dagger
   \or \ddagger\ddagger \else\@ctrerr\fi}}

\begin{document}
\pagestyle{headings}
\mainmatter
\def\ECCVSubNumber{3852}  

\makeatletter
\def\@fnsymbol#1{\ensuremath{\ifcase#1\or *\or \dagger\or \ddagger\or
   \mathsection\or \mathparagraph\or \|\or **\or \dagger\dagger
   \or \ddagger\ddagger \else\@ctrerr\fi}}
    \makeatother

\title{\modelname: DETR with Improved DeNoising Anchor Boxes for End-to-End Object Detection} 
\titlerunning{DINO: DETR with Improved DeNoising Anchor Boxes}
\authorrunning{H. Zhang, F. Li, S. Liu, et al.} 
\author{Hao Zhang$^{1,3}$\thanks{Equal contribution. Listing order is random.}\thanks{This work was done when Hao Zhang, Feng Li, and Shilong Liu were interns at IDEA. }, ~Feng Li$^{1,3*\dag}$, ~Shilong Liu$^{2,3*\dag}$, ~Lei Zhang$^{3}$\thanks{Corresponding author.},\\ ~Hang Su$^{2}$, ~Jun Zhu$^{2}$, ~Lionel M. Ni$^{1,4}$, ~Heung-Yeung Shum$^{1,3}$ \\
}
\institute{$^1$The Hong Kong University of Science and Technology. \\
$^2$Dept. of CST., BNRist Center, Institute for AI, Tsinghua University. \\
$^3$International Digital Economy Academy (IDEA). \\
$^4$The Hong Kong University of Science and Technology (Guangzhou).\\\
\texttt{\{hzhangcx,fliay\}@connect.ust.hk} \\
\texttt{\{liusl20\}@mails.tsinghua.edu.cn} \\
\texttt{\{suhangss,dcszj\}@mail.tsinghua.edu.cn} \\
\texttt{\{ni,hshum\}@ust.hk} \\
\texttt{\{leizhang\}@idea.edu.cn} \\
}

\maketitle


\begin{abstract}
We present  {\modelname} (\textbf{D}ETR with \textbf{I}mproved de\textbf{N}oising anch\textbf{O}r boxes), a state-of-the-art end-to-end object detector. 
DINO improves over previous DETR-like models in performance and efficiency by using a contrastive way for denoising training, a mixed query selection method for anchor initialization, and a look forward twice scheme for box prediction.
DINO achieves $49.4$AP in $12$ epochs and $51.3$AP in $24$ epochs on COCO with a ResNet-50 backbone and multi-scale features, yielding a significant improvement of $\textbf{+6.0}$\textbf{AP} and $\textbf{+2.7}$\textbf{AP}, respectively, compared to DN-DETR, the previous best DETR-like model. DINO scales well in both model size and data size. Without bells and whistles, after pre-training on the Objects365 dataset with a SwinL backbone, DINO obtains the best results on both COCO \texttt{val2017} ($\textbf{63.2}$\textbf{AP}) and \texttt{test-dev} (\textbf{$\textbf{63.3}$AP}). Compared to other models on the leaderboard, DINO significantly reduces its model size and pre-training data size while achieving better results. Our code will be available at \url{https://github.com/IDEACVR/DINO}.

\end{abstract}

\begin{figure}[ht]
    \centering
    \begin{center}
\includegraphics[width=.44\columnwidth]{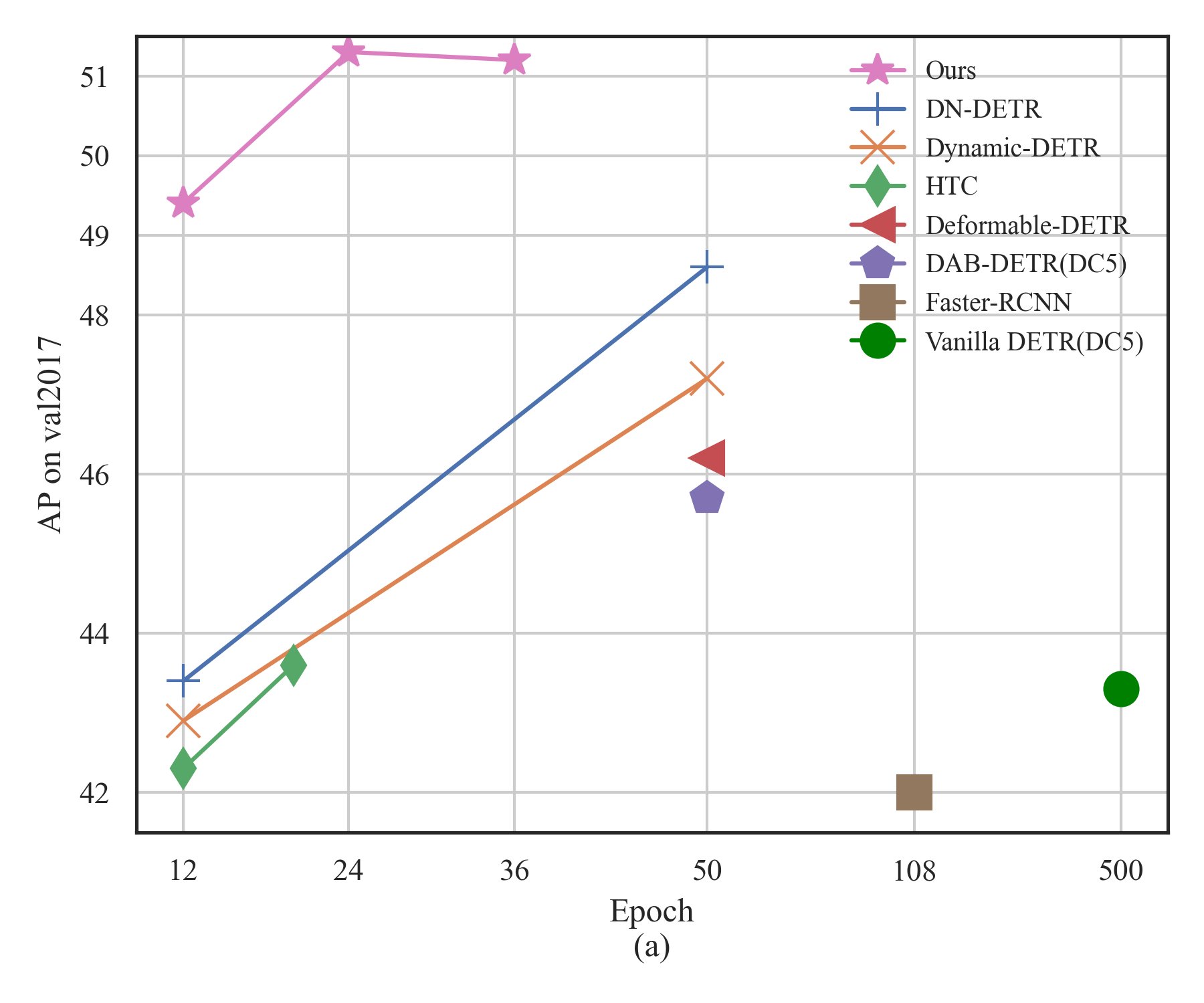}
\includegraphics[width=.42\columnwidth]{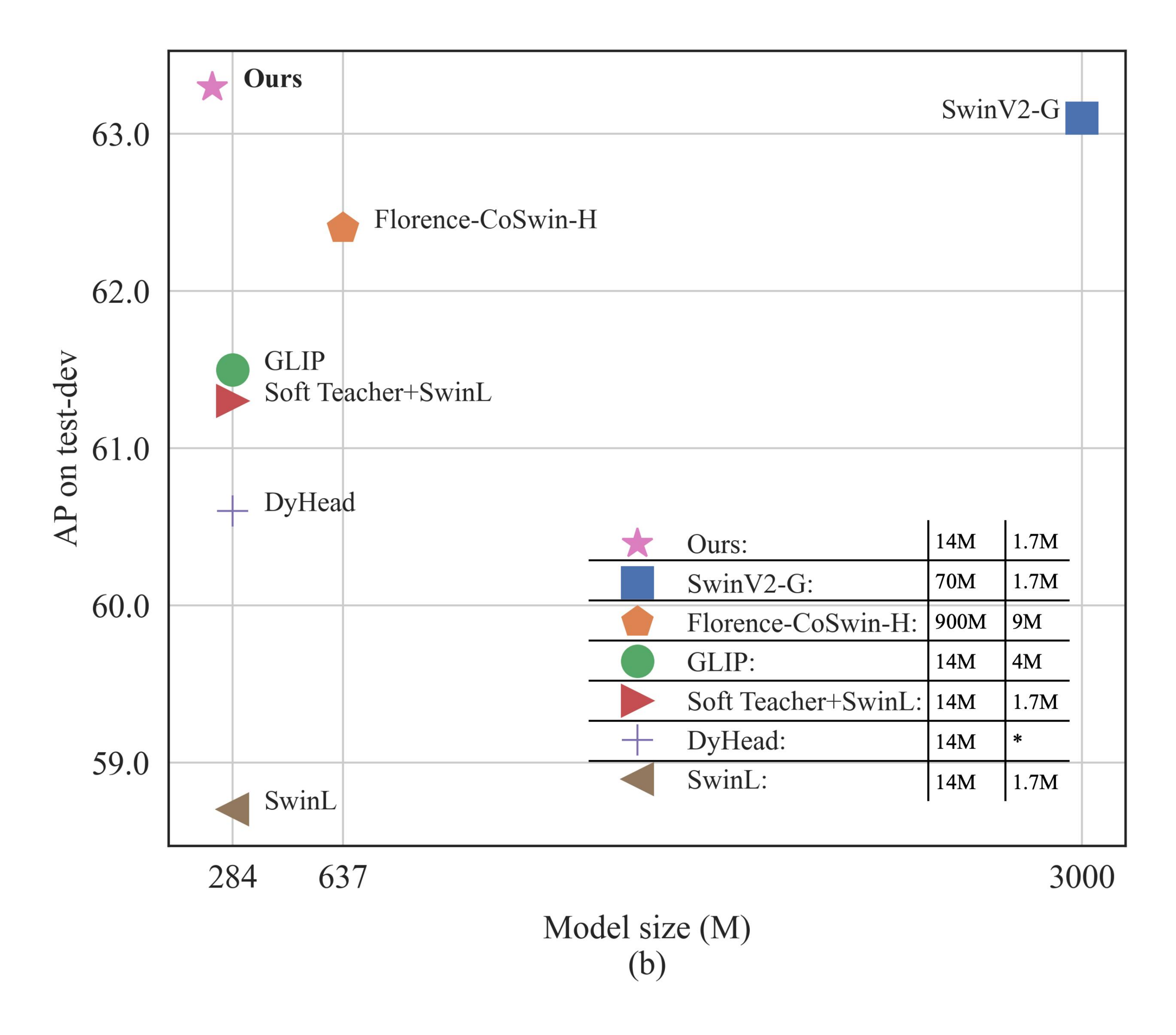}
\vspace{-0.4cm}
\end{center}
    \caption{AP on COCO compared with other detection models. (a) Comparison to models with a ResNet-50 backbone w.r.t. training epochs. Models marked with DC5 use a dilated larger resolution feature map. Other models use multi-scale features. (b) Comparison to SOTA models w.r.t. pre-training data size and model size. SOTA models are from the COCO  \texttt{test-dev} leaderboard. In the legend we list the backbone pre-training data size (first number) and detection pre-training data size (second number). $*$ means the data size is not disclosed.
    }
    \label{fig:intro}
\end{figure}

\section{Introduction}
Object detection is a fundamental task in computer vision. Remarkable progress has been accomplished by classical convolution-based object detection algorithms~\cite{RenHG017,tian2019fcos,lin2018focal,bochkovskiy2020yolov4,ge2021yolox}. Despite that such algorithms normally include hand-designed components like anchor generation and non-maximum suppression (NMS), they yield the best detection models such as DyHead~\cite{dai2021dynamic}, Swin~\cite{liu2021swin} and SwinV2~\cite{liu2021swinv2} with HTC++~\cite{chen2019hybrid}, as evidenced on the COCO test-dev leaderboard \cite{paperwithcode}.

In contrast to classical detection algorithms, DETR~\cite{carion2020end} is a novel Transformer-based detection algorithm. It eliminates the need of hand-designed components and achieves comparable performance with optimized classical detectors like Faster RCNN~\cite{RenHG017}. Different from previous detectors, DETR models object detection as a set prediction task and assigns labels by bipartite graph matching. It leverages learnable queries to probe the existence of objects and combine features from an image feature map, which behaves like soft ROI pooling~\cite{liu2022dab}.

Despite its promising performance, the training convergence of DETR is slow and the meaning of queries is unclear. To address such problems, many methods have been proposed, such as introducing deformable attention~\cite{zhu2020deformable}, decoupling positional and content information~\cite{meng2021conditional}, providing spatial priors~\cite{gao2021fast,yao2021efficient,wang2021anchor}, etc. Recently, DAB-DETR~\cite{liu2022dab} proposes to formulate DETR queries as dynamic anchor boxes (DAB), which bridges the gap between classical anchor-based detectors and DETR-like ones. DN-DETR~\cite{li2022dn} further solves the instability of bipartite matching by introducing a denoising (DN) technique. The combination of DAB and DN makes DETR-like models competitive with classical detectors on both training efficiency and inference performance.

The best detection models nowadays are based on improved classical detectors like DyHead~\cite{Dai_2021_ICCV} and HTC~\cite{chen2019hybrid}. For example, the best result presented in SwinV2~\cite{liu2021swinv2} was trained with the HTC++~\cite{chen2019hybrid,liu2021swin} framework. 
Two main reasons contribute to the phenomenon: 1) 
\textit{Previous DETR-like models are inferior} to the improved classical detectors. Most classical detectors have been well studied and highly optimized, leading to a better performance compared with the newly developed DETR-like models. For instance, the best performing DETR-like models nowadays are still under $50$ AP on COCO.  
2) The \textit{scalability} of DETR-like models has not been well studied. There is no reported result about how DETR-like models perform when scaling to a large backbone and a large-scale data set. We aim to address both concerns in this paper.

Specifically, by improving the denoising training, query initialization, and box prediction, we 
design a new DETR-like model based on DN-DETR~\cite{li2022dn},  DAB-DETR~\cite{liu2022dab}, and Deformable DETR~\cite{zhu2020deformable}. 
We name our model as \textbf{\modelname} (\textbf{D}ETR with \textbf{I}mproved de\textbf{N}oising anch\textbf{O}r box).
As shown in Fig. \ref{fig:intro}, the comparison on COCO shows the superior performance of {\modelname}.
In particular, {\modelname} demonstrates a great scalability, setting a new record of $63.3$ AP on the COCO test-dev leaderboard \cite{paperwithcode}

As a DETR-like model, {\modelname} contains a backbone, a multi-layer Transformer encoder, a multi-layer Transformer decoder, and multiple prediction heads. Following DAB-DETR~\cite{liu2022dab}, we formulate queries in decoder as dynamic anchor boxes and refine them step-by-step across decoder layers. Following DN-DETR~\cite{li2022dn}, we add ground truth labels and boxes with noises into the Transformer decoder layers to help stabilize bipartite matching during training. We also adopt deformable attention~\cite{zhu2020deformable} for its computational efficiency. Moreover, we propose three new methods as follows. 
First, to improve the one-to-one matching, we propose a \textit{contrastive denoising training} by adding both positive and negative samples of the same ground truth at the same time. After adding two different noises to the same ground truth box, we mark the box with a smaller noise as positive and the other as negative. The contrastive denoising training helps the model to avoid duplicate outputs of the same target. 
Second, the dynamic anchor box formulation of queries links DETR-like models with classical two-stage models. Hence we propose a \textit{mixed query selection} method, which helps better initialize the queries. We select initial anchor boxes as positional queries from the output of the encoder, similar to~\cite{zhu2020deformable,yao2021efficient}. However, we leave the content queries learnable as before, encouraging the first decoder layer to focus on the spatial prior. 
Third, to leverage the refined box information from later layers to help optimize the parameters of their adjacent early layers,
we propose a new \textit{look forward twice} scheme to correct the updated parameters with gradients from later layers.

We validate the effectiveness of DINO with extensive experiments on the COCO~\cite{lin2015microsoft} detection benchmarks. As shown in Fig. \ref{fig:intro}, DINO achieves $49.4$AP in $12$ epochs and $51.3$AP in $24$ epochs with ResNet-50 and multi-scale features, yielding a significant improvement of $\textbf{+6.0}$AP and $\textbf{+2.7}$AP, respectively, compared to the previous best DETR-like model.
In addition, DINO scales well in both model size and data size. After pre-training on the Objects365~\cite{shao2019objects365} data set with a  SwinL~\cite{liu2021swin} backbone, DINO achieves the best results on both COCO \texttt{val2017} ($\textbf{63.2}$AP) and \texttt{test-dev} ($\textbf{63.3}$AP) benchmarks, as shown in Table \ref{tab:sota}.
Compared to other models on the leaderboard \cite{paperwithcode}, we reduce the model size to $\textbf{1/15}$ compared to SwinV2-G~\cite{liu2021swinv2}. Compared to Florence~\cite{yuan2021florence}, we reduce the pre-training detection dataset to $\textbf{1/5}$ and backbone pre-training dataset to $\textbf{1/60}$ while achieving better results.

We summarize our contributions as follows.
\vspace{-0.1cm}
\begin{enumerate}
    \item We design a new end-to-end DETR-like object detector with several novel techniques, including 
    contrastive DN training, mixed query selection, and look forward twice for different parts of the DINO model. 
    \item We conduct intensive ablation studies to validate the effectiveness of different design choices in DINO. As a result, DINO achieves $49.4$AP in $12$ epochs and $51.3$AP in $24$ epochs with ResNet-50 and multi-scale features, significantly outperforming the previous best DETR-like models. In particular, DINO trained in $12$ epochs shows a more significant improvement on small objects, yielding an improvement of $\textbf{+7.5}$AP.
    \item We show that, without bells and whistles, DINO can achieve the best performance on public benchmarks. After pre-training on the Objects365~\cite{shao2019objects365} dataset with a SwinL~\cite{liu2021swin} backbone, DINO achieves the best results on both COCO \texttt{val2017} ($\textbf{63.2}$AP) and \texttt{test-dev} ($\textbf{63.3}$AP) benchmarks. To the best of our knowledge, this is the first time that an end-to-end Transformer detector outperforms state-of-the-art (SOTA) models on the COCO leaderboard \cite{paperwithcode}.
\end{enumerate}

\section{Related Work}
\subsection{Classical Object Detectors}
Early convolution-based object detectors are either two-stage or one-stage models, based on hand-crafted anchors or reference points. Two-stage models~\cite{ren2015faster,he2017mask} usually use an region proposal network (RPN)~\cite{ren2015faster} to propose potential boxes, which are then refined in the second stage. One-stage models such as YOLO v2~\cite{redmon2017yolo9000} and YOLO v3~\cite{redmon2018yolov3} directly output offsets relative to predefined anchors. Recently, some convolution-based models such as HTC++~\cite{chen2019hybrid} and Dyhead~\cite{dai2021dynamic} have achieved top performance on the COCO 2017 dataset~\cite{lin2015microsoft}. 
The performance of convolution-based models, however, relies on the way they generate anchors. Moreover, they need hand-designed components like NMS to remove duplicate boxes, and hence cannot perform end-to-end optimization.

\subsection{DETR and Its Variants}
Carion \textit{et al.} \cite{carion2020end} proposed a Transformer-based end-to-end object detector named DETR (DEtection TRansformer) without using hand-designed components like anchor design and NMS. Many follow-up papers have attempted to address the slow training convergence issue of DETR introduced by decoder cross-attention. For instance, Sun \textit{et al.} \cite{sun2020rethinking} designed an encoder-only DETR without using a decoder. Dai \textit{et al.} \cite{dai2021dynamic} proposed a dynamic decoder to focus on important regions from multiple feature levels.

Another line of works is towards a deeper understanding of decoder queries in DETR. Many papers associate queries with spatial position from different perspectives. Deformable DETR~\cite{zhu2020deformable} predicts $2$D anchor points and designs a deformable attention module that only attends to certain sampling points around a reference point. Efficient DETR~\cite{yao2021efficient} selects top K positions from encoder's dense prediction to enhance decoder queries. DAB-DETR~\cite{liu2022dab} further extends $2$D anchor points to $4$D anchor box coordinates to represent queries and dynamically update boxes in each decoder layer. Recently, DN-DETR~\cite{li2022dn} introduces a denoising training method to speed up DETR training. It feeds noise-added ground-truth labels and boxes into the decoder and trains the model to reconstruct the original ones. Our work of DINO in this paper is based on DAB-DETR and DN-DETR, and also adopts deformable attention for its computational efficiency.

\subsection{Large-scale Pre-training for Object Detection}
Large-scale pre-training have had a big impact on both natural language processing~\cite{devlin2018bert} and computer vision~\cite{AlecRadford2021LearningTV}. The best performance detectors nowadays are mostly achieved with large backbones pre-trained on large-scale data. For example, Swin V2~\cite{liu2021swinv2} extends its backbone size to $3.0$ Billion parameters
and pre-trains its models with $70$M privately collected images.
Florence~\cite{yuan2021florence} first pre-trains its backbone with $900$M privately curated image-text pairs and then pre-trains its detector with $9$M images with annotated or pseudo boxes. In contrast, DINO achieves the SOTA result with a publicly available SwinL~\cite{liu2021swin} backbone and a public dataset Objects365~\cite{shao2019objects365} (1.7M annotated images) only.

\section{\modelname: {D}ETR with {I}mproved De{N}oising Anchor Boxes}
\label{sec:method}
\begin{figure}[h]
    \includegraphics[width=\columnwidth]{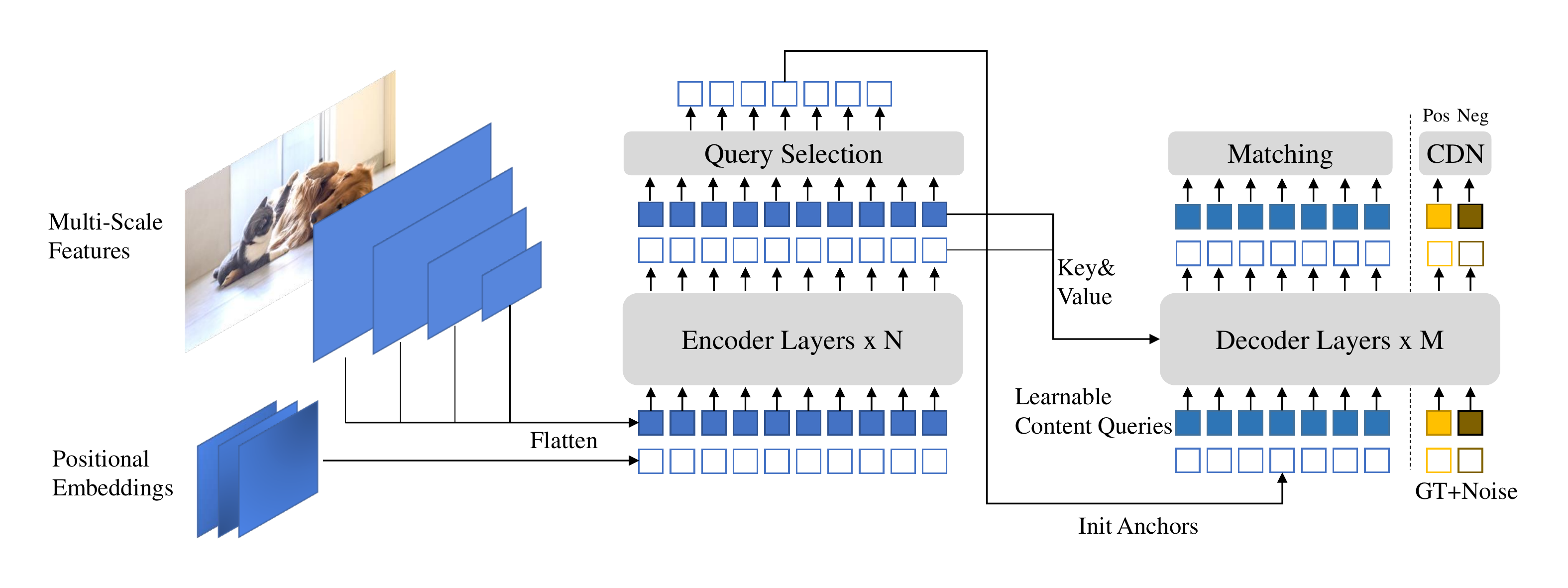}
    \centering
    \vspace{-0.5cm}
    \caption{The framework of our proposed DINO model. Our improvements are mainly in the Transformer encoder and decoder. The top-K encoder features in the last layer are selected to initialize the positional queries for the Transformer decoder, whereas the content queries are kept as learnable parameters. Our decoder also contains a Contrastive DeNoising (CDN) part with both positive and negative samples. }
    \label{fig:framework}
\end{figure}
\subsection{Preliminaries}\label{sec:preliminary}
As studied in Conditional DETR~\cite{meng2021conditional} and DAB-DETR~\cite{liu2022dab}, it becomes clear that queries in DETR~\cite{carion2020end} are formed by two parts: a positional part and a content part, which are referred to as positional queries and content queries in this paper. 
DAB-DETR~\cite{liu2022dab} explicitly formulates each positional query in DETR as a 4D anchor box $(x,y,w,h)$, where $x$ and $y$ are the center coordinates of the box and $w$ and $h$ correspond to its width and height. Such an explicit anchor box formulation makes it easy to dynamically refine anchor boxes layer by layer in the decoder.

DN-DETR~\cite{li2022dn} introduces a denoising (DN) training method to accelerate the training convergence of DETR-like models. It shows that the slow convergence problem in DETR is caused by the instability of bipartite matching. To mitigate this problem, DN-DETR proposes to additionally feed noised ground-truth (GT) labels and boxes into the Transformer decoder and train the model to reconstruct the ground-truth ones. The added noise $(\Delta x, \Delta y, \Delta w, \Delta h)$ is constrained by $|\Delta x|<\frac{\lambda w}{2}$, $|\Delta y|<\frac{\lambda h}{2}$, $|\Delta w|<\lambda w$, and $|\Delta y|<\lambda h$, where $(x,y,w,h)$ denotes a GT box and $\lambda$\footnote{The DN-DETR paper~\cite{li2022dn} uses $\lambda_1$ and $\lambda_2$ to denote noise scales of center shifting and box scaling, but sets $\lambda_1 =\lambda_2$. In this paper, we use $\lambda$ in place of $\lambda_1 \text{and} \lambda_2$ for simplicity.} is a hyper-parameter to control the scale of noise. Since DN-DETR follows DAB-DETR to view decoder queries as anchors, a noised GT box can be viewed as a special anchor with a GT box nearby as $\lambda$ is usually small. In addition to the orginal DETR queries, DN-DETR adds a DN part which feeds noised GT labels and boxes into the decoder to provide an auxiliary DN loss. The DN loss effectively stabilizes and speeds up the DETR training and can be plugged into any DETR-like models. 

Deformable DETR~\cite{zhu2020deformable} is another early work to speed up the convergence of DETR. To compute deformable attention, it introduces the concept of reference point so that deformable attention can attend to a small set of key sampling points around a reference. The reference point concept makes it possible to develop several techniques to further improve the DETR performance. The first technique is query selection\footnote{Also named as ``two-stage'' in the Deformable DETR paper. As the ``two-stage'' name might confuse readers with classical detectors, we use the term ``query selection'' instead in our paper.}, which selects features and reference boxes from the encoder as inputs to the decoder directly. The second technique is iterative bounding box refinement with a careful gradient detachment design between two decoder layers. We call this gradient detachment technique ``look forward once'' in our paper. 

Following DAB-DETR and DN-DETR, {\modelname} formulates the positional queries as dynamic anchor boxes and is trained with an extra DN loss. Note that DN-DETR
also adopts several techniques from Deformable DETR to achieve a better performance, including its deformable attention mechanism and ``look forward once'' implementation in layer parameter update. DINO further adopts the query selection idea from Deformable DETR to better initialize the positional queries. Built upon this strong baseline, {\modelname} introduces three novel methods to further improve the detection performance, which will be described in Sec. \ref{sec:contrastive_dn}, Sec. \ref{sec:mixed_query_selection}, and Sec. \ref{sec:look_forward}, respectively.

\subsection{Model Overview}
As a DETR-like model, DINO is an end-to-end architecture which contains a backbone, a multi-layer Transformer~\cite{vaswani2017attention} encoder, a multi-layer Transformer decoder, and multiple prediction heads. The overall pipeline is shown in Fig. \ref{fig:framework}. 
Given an image, we extract multi-scale features with backbones like ResNet~\cite{he2015deep} or Swin Transformer~\cite{liu2021swin}, and then feed them into the Transformer encoder with corresponding positional embeddings. After feature enhancement with the encoder layers, we propose a new mixed query selection strategy to initialize anchors as positional queries for the decoder. Note that this strategy does not initialize content queries but leaves them learnable. More details of mixed query selection are available in Sec. \ref{sec:mixed_query_selection}. 
With the initialized anchors and the learnable content queries, we use the deformable attention~\cite{zhu2020deformable} to combine the features of the encoder outputs and update the queries layer-by-layer. The final outputs are formed with refined anchor boxes and classification results predicted by refined content features. 
As in DN-DETR~\cite{li2022dn}, we have an extra DN branch to perform denoising training. Beyond the standard DN method, we propose a new contrastive denoising training approach by taking into account hard negative samples, which will be presented in Sec. \ref{sec:contrastive_dn}. 
To fully leverage the refined box information from later layers to help optimize the parameters of their adjacent early layer, a novel look forward twice method is proposed to pass gradients between adjacent layers, which will be described in Sec. \ref{sec:look_forward}.


\subsection{Contrastive DeNoising Training}
\label{sec:contrastive_dn}

\begin{figure}[h]
    \includegraphics[width=0.8\columnwidth]{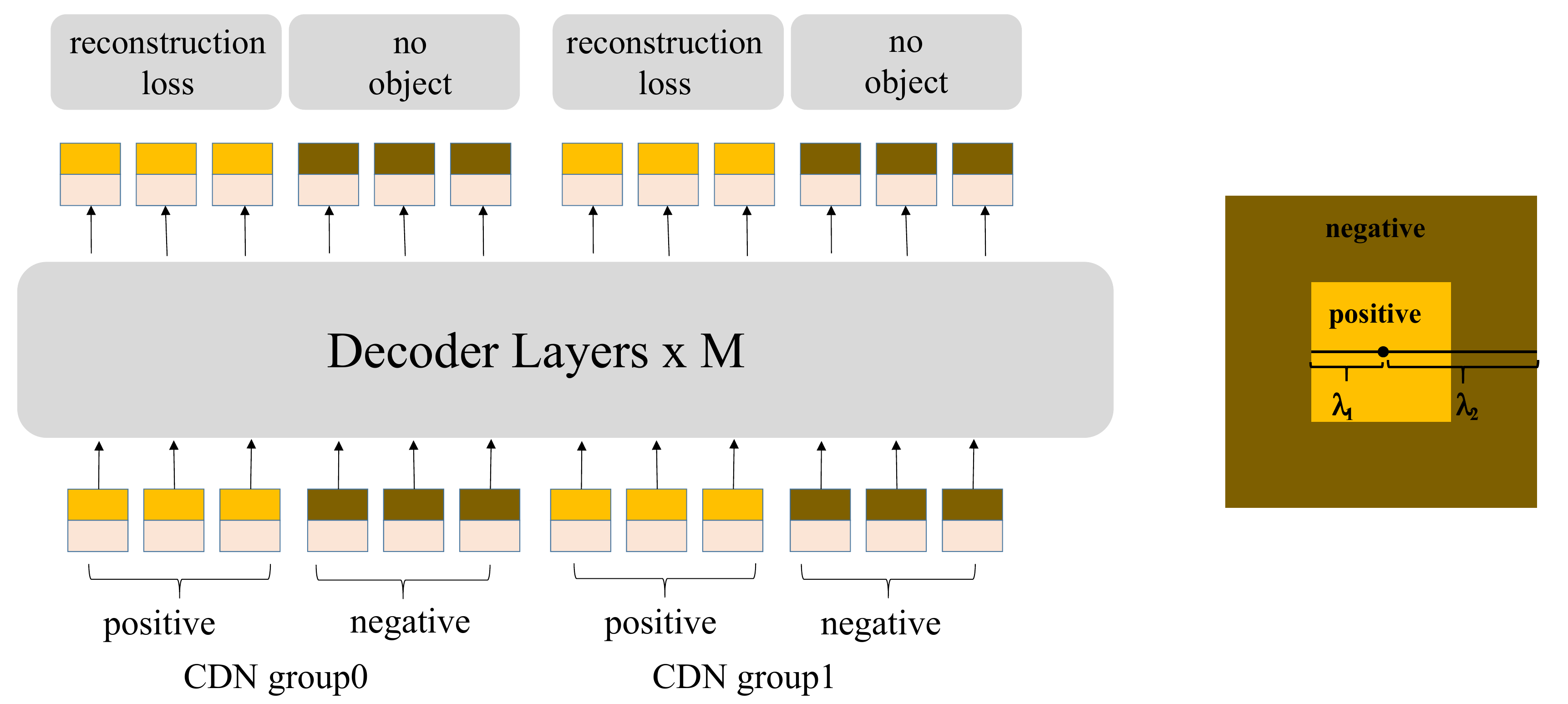}
    \centering
    \vspace{-0.3cm}
    \caption{The structure of CDN group and a demonstration of positive and negative examples.
    Although both positive and negative examples are $4$D anchors that can be represented as points in $4$D space, we illustrate them as points in $2$D space on concentric squares for simplicity. Assuming the square center is a GT box, points inside the inner square are regarded as a positive example and points between the inner square and the outer square are viewed as negative examples.
    }
    \label{fig:CDN}
\end{figure}
DN-DETR is very effective in stabilizing training and accelerating convergence. With the help of DN queries, it learns to make predictions based on anchors which have GT boxes nearby. However, it lacks a capability of predicting ``no object" for anchors with no object nearby. To address this issue, we propose a Contrastive DeNoising (CDN) approach to \emph{rejecting} useless anchors. 

\noindent\textbf{Implementation:} DN-DETR has a hyper-parameter $\lambda$ to control the noise scale. The generated noises are no larger than $\lambda$ as DN-DETR wants the model to reconstruct the ground truth (GT) from moderately noised queries. In our method, we have two hyper-parameters $\lambda_1$ and $\lambda_2$, where $\lambda_1 < \lambda_2$. As shown in the concentric squares in Fig.~\ref{fig:CDN}, we generate two types of CDN queries: positive queries and negative queries. Positive queries within the inner square have a noise scale smaller than $\lambda_1$ and are expected to reconstruct their corresponding ground truth boxes. Negative queries between the inner and outer squares have a noise scale larger than $\lambda_1$ and smaller than $\lambda_2$. They are are expected to predict ``no object". We usually adopt small $\lambda_2$ because hard negative samples closer to GT boxes are more helpful to improve the performance. As shown in Fig.~\ref{fig:CDN}, each CDN group has a set of positive queries and negative queries. If an image has $n$ GT boxes, a CDN group will have $2\times n$ queries with each GT box generating a positive and a negative queries. Similar to DN-DETR, we also use multiple CDN groups to improve the effectiveness of our method. The reconstruction losses are $l_1$ and GIOU losses for 
box regression and focal loss~\cite{lin2018focal} for classification. The loss to classify negative samples as background is also focal loss.

\begin{figure}[h]
    \centering
    \begin{minipage}[t]{0.45\textwidth}
    \begin{center}
\includegraphics[width=5cm]{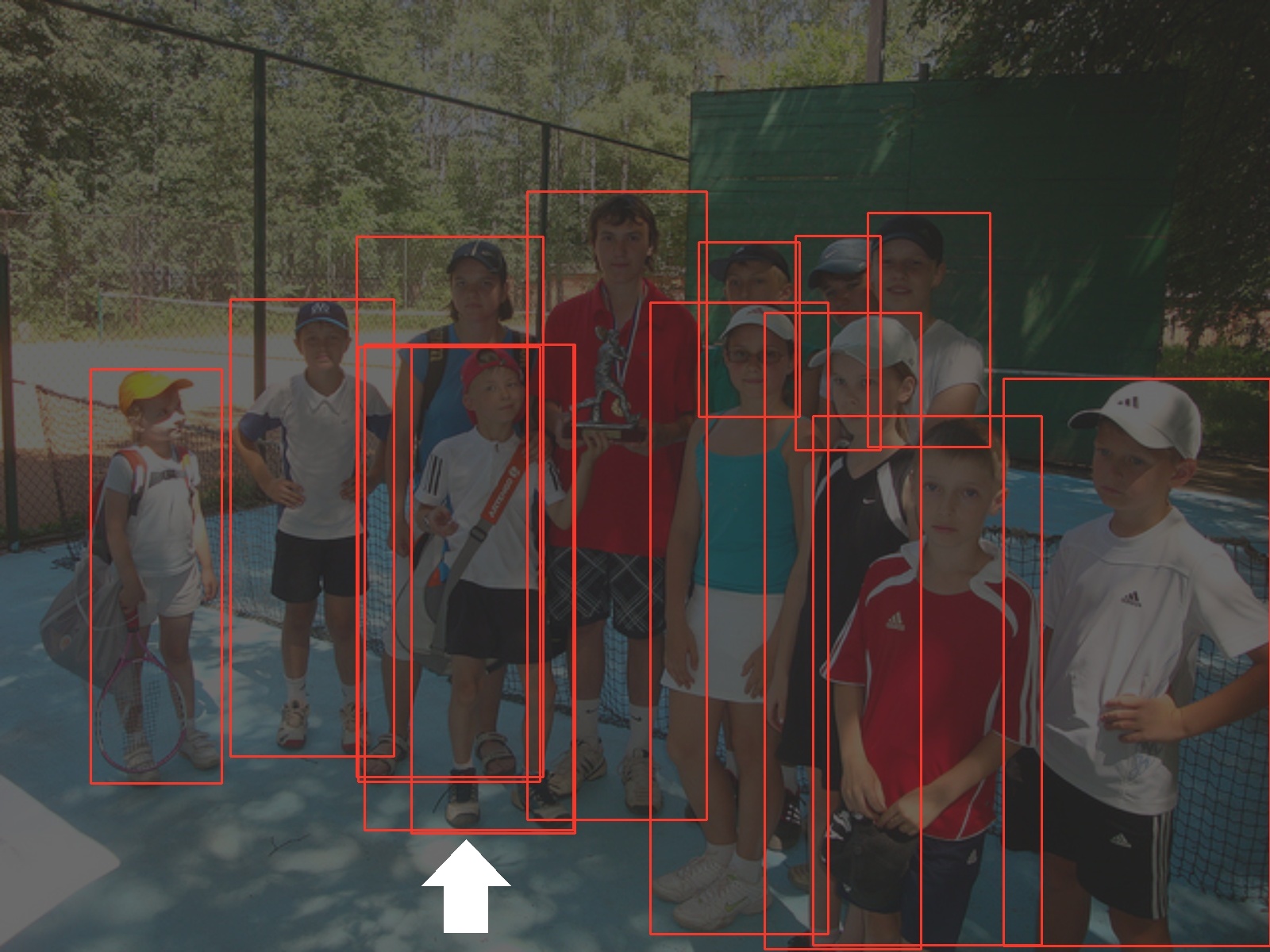}
\end{center}
\end{minipage}
\hfill
\vspace{-0.3cm}
\begin{minipage}[t]{0.45\textwidth}
\begin{center}
\includegraphics[width=5cm]{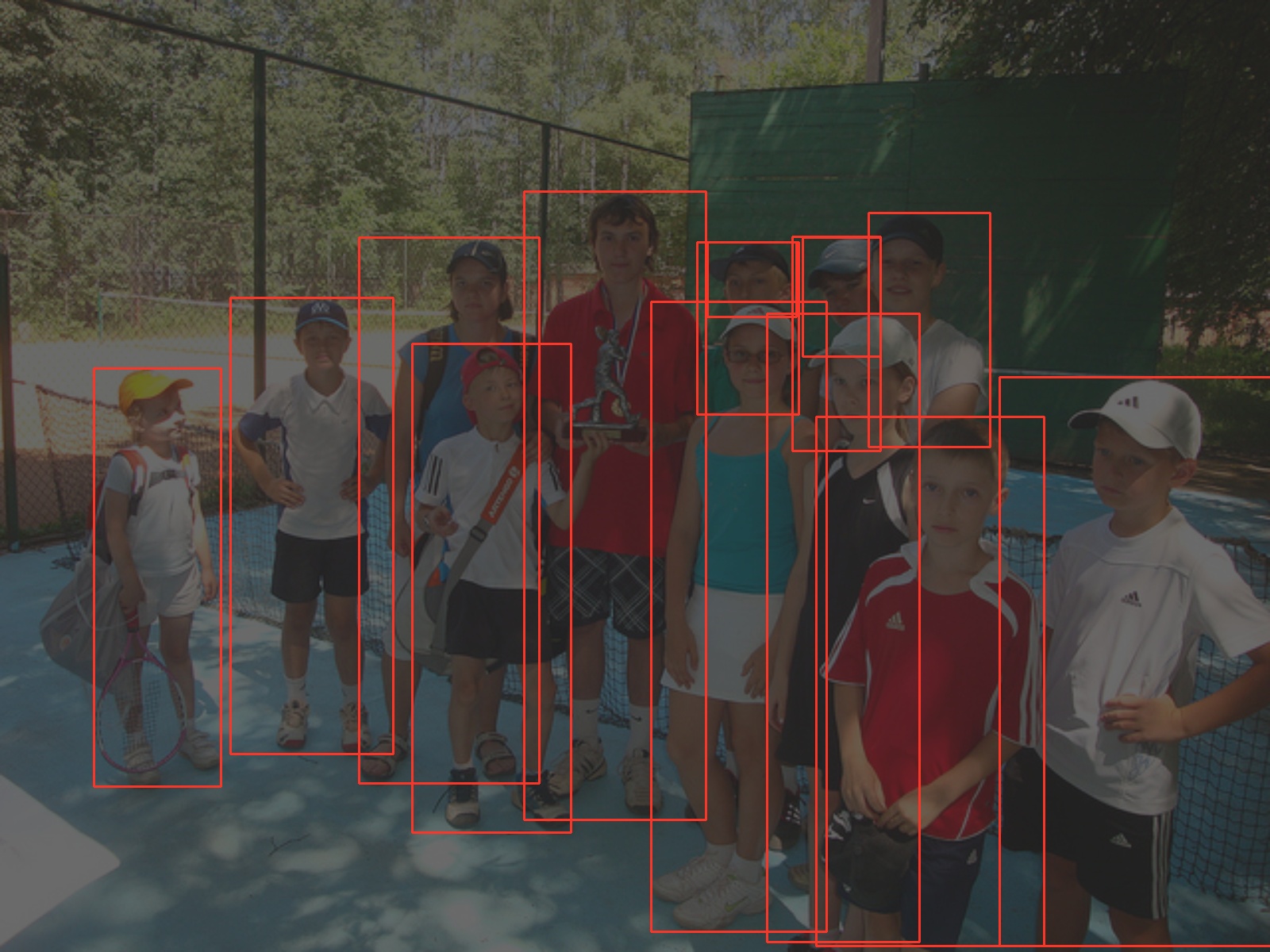}
\vspace{-0.2cm}
\end{center}
\end{minipage}

    \caption{The left figure is the detection result using a model trained with DN queries and the right is the result of our method. In the left image, the boy pointed by the arrow has $3$ duplicate bounding boxes. For clarity, we only show boxes of class ``person".}
\label{fig:sample images}
    \vspace{-0.3cm}
\end{figure}

\noindent\textbf{Analysis:} The reason why our method works is that it can inhibit confusion and select high-quality anchors (queries) for predicting bounding boxes. The confusion happens when multiple anchors are close to one object. In this case, it is hard for the model to decide which anchor to choose. The confusion may lead to two problems. The first is duplicate predictions. Although DETR-like models can inhibit duplicate boxes with the help of set-based loss and self-attention~\cite{carion2020end}, this ability is limited. As shown in the left figure of Fig.~\ref{fig:sample images}, when replacing our CDN queries with DN queries, the boy pointed by the arrow has $3$ duplicate predictions. With CDN queries, our model can distinguish the slight difference between anchors and avoid duplicate predictions as shown in the right figure of Fig.~\ref{fig:sample images}. The second problem is that an unwanted anchor farther from a GT box might be selected. Although denoising training~\cite{li2022dn} has improved the model to choose nearby anchors, CDN further improves this capability by teaching the model to reject farther anchors.

\begin{figure}[htbp]
\vspace{-0.3cm}
\centering
\begin{minipage}[t]{0.3\textwidth}
\centering
\includegraphics[width=4cm]{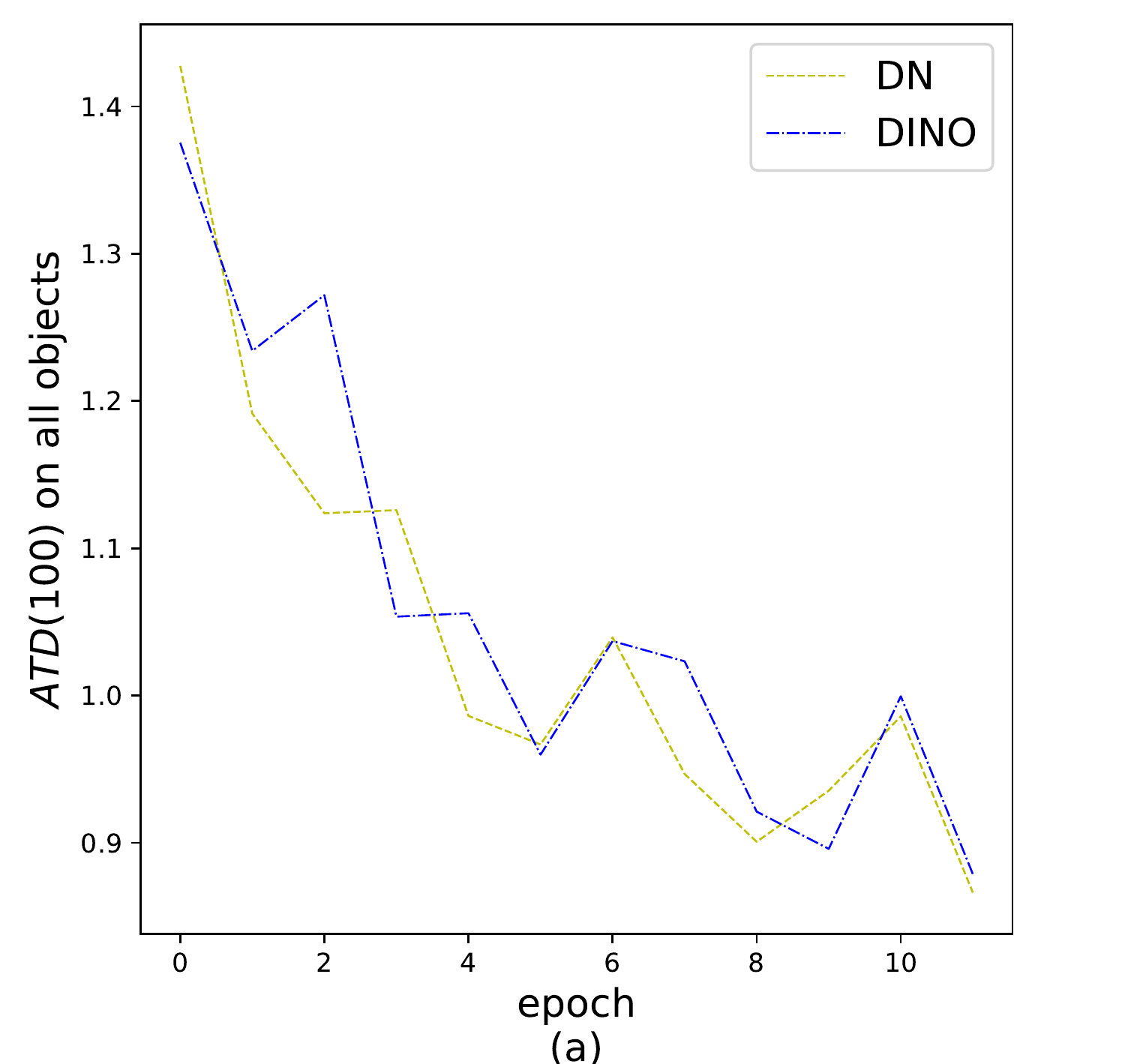}
\end{minipage}
\hfill
\begin{minipage}[t]{0.3\textwidth}
\centering
\includegraphics[width=4cm]{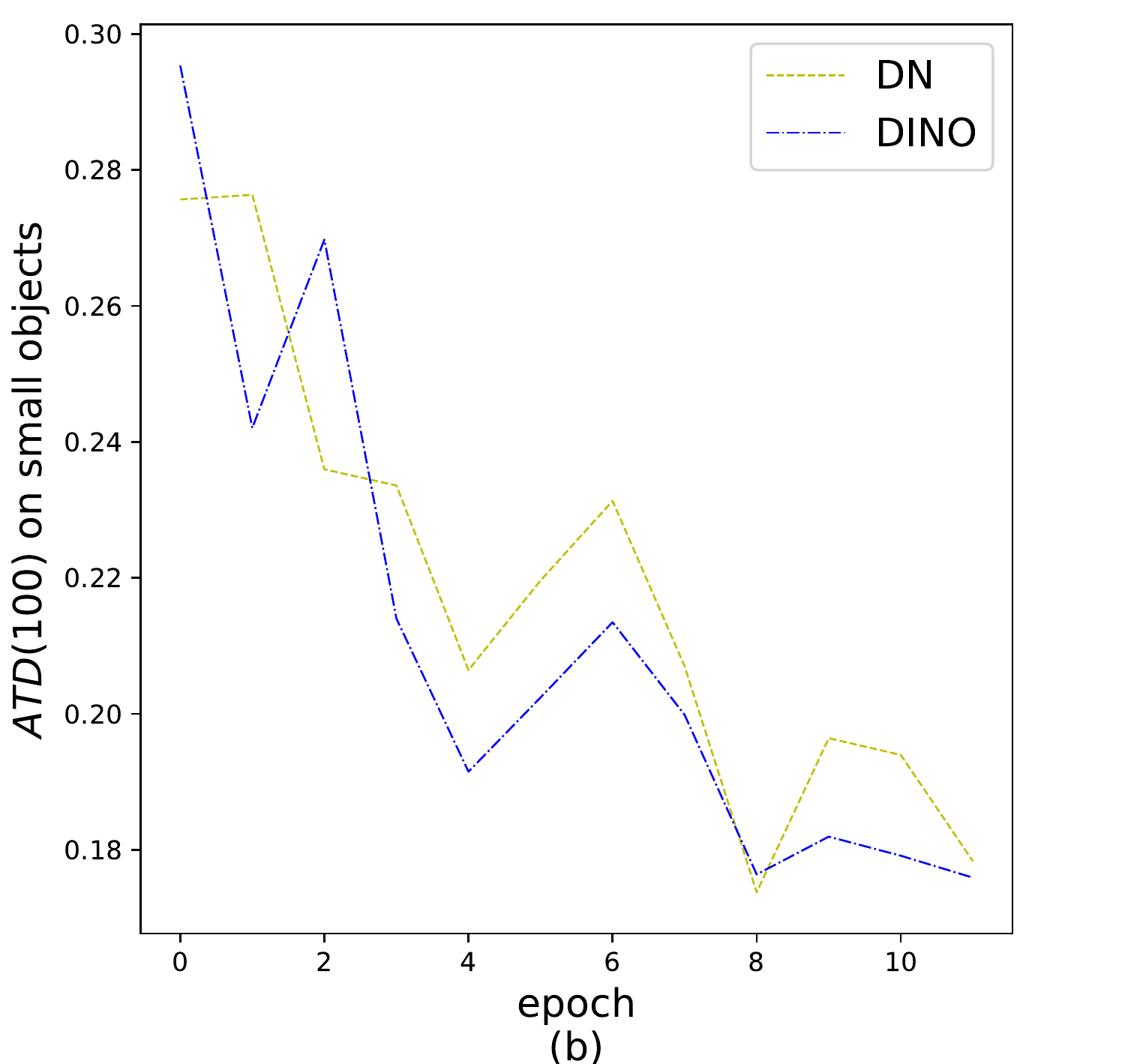}
\end{minipage}
\hfill
\begin{minipage}[t]{0.3\textwidth}
\centering
\includegraphics[width=4cm]{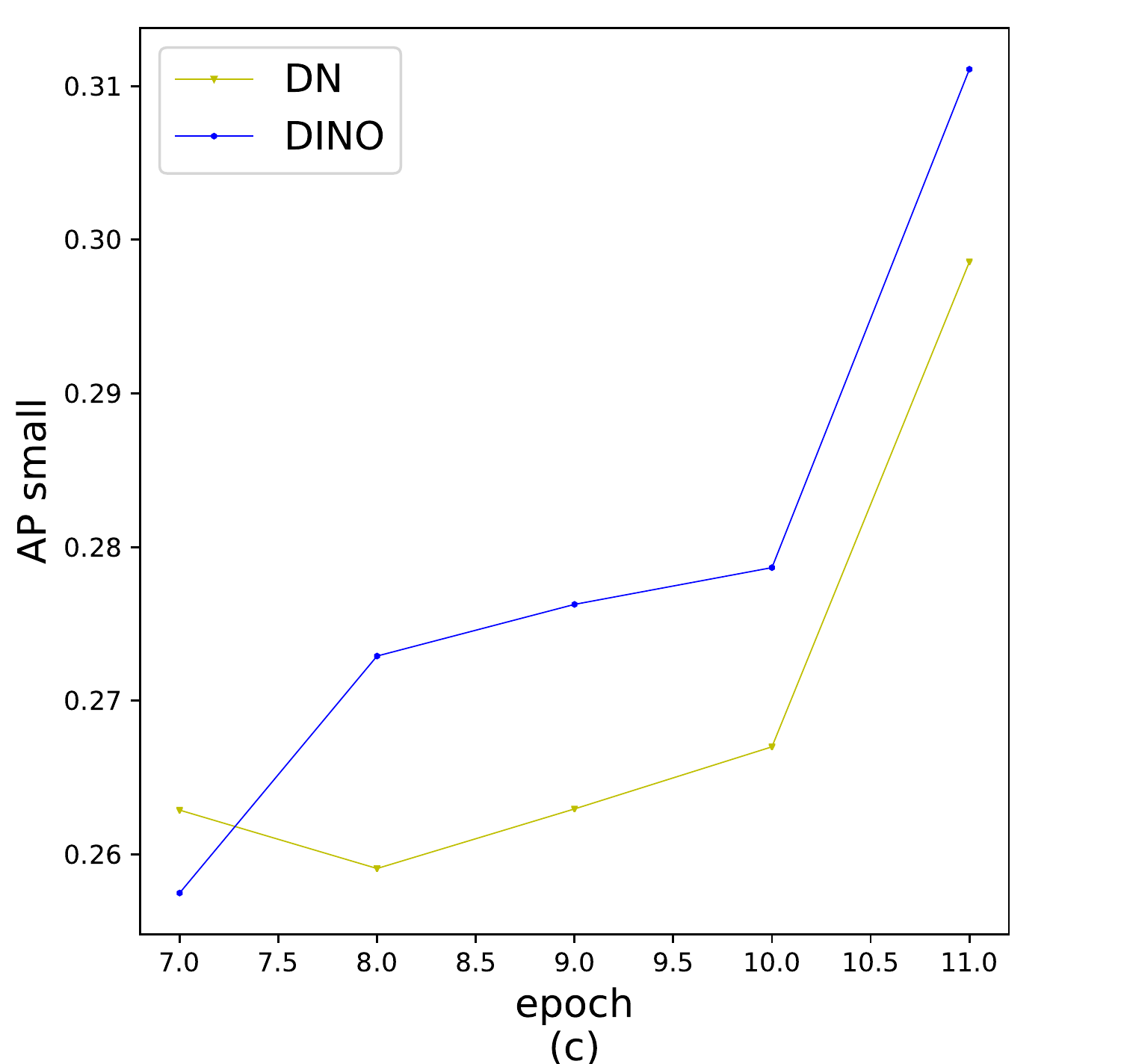}
\end{minipage}
\vspace{-0.3cm}
\caption{(a) and (b) $ATD(100)$ on all objects and small objects respectively. (c) The AP on small objects.}
\label{fig:neg_dist}
\vspace{-.1cm}
\end{figure}

\noindent\textbf{Effectiveness:} To demonstrate the effectiveness of CDN, we define a metric called Average Top-K Distance (ATD($k$)) and use it to evaluate how far anchors are from their target GT boxes in the matching part.
As in DETR, each anchor corresponds to a prediction which may be matched with a GT box or background. We only consider those matched with GT boxes here. Assume we have $N$ GT bounding boxes ${b_0, b_2, ..., b_{N-1}}$ in a validation set, where $b_i=(x_i, y_i, w_i, h_i)$. For each $b_i$, we can find its corresponding anchor and denote it as $a_i=(x^{\prime}_i, y^{\prime}_i, w^{\prime}_i, h^{\prime}_i)$. $a_i$ is the initial anchor box of the decoder whose refined box after the last decoder layer is assigned to $b_i$ during matching. Then we have 
\begin{equation}
    ATD(k)=\frac{1}{k}\sum\left\{\mathop{topK}\left(\left\{\lVert b_0-a_0 \rVert_{1}, \lVert b_1-a_1 \rVert_{1}, ..., \lVert b_{N-1}-a_{N-1} \rVert_{1}\right\},k\right)\right\}
\end{equation}
where $\lVert b_i-a_i \rVert_{1}$ is the $l_1$ distance between $b_i$ and $a_i$ and $\mathop{topK}(\mathbf{x},k)$ is a function that returns the set of $k$ largest elements in $\mathbf{x}$. The reason why we select the top-K elements is that the confusion problem is more likely to happen when the GT box is matched with a farther anchor. As shown in (a) and (b) of Fig.~\ref{fig:neg_dist}, DN is good enough for selecting a good anchor overall. However, CDN finds better anchors for small objects. Fig.~\ref{fig:neg_dist} (c) shows that CDN queries lead to an improvement of $+1.3$ AP over DN queries on small objects in $12$ epochs with ResNet-50 and multi-scale features.

\subsection{Mixed Query Selection}
\label{sec:mixed_query_selection}

\begin{figure}[h]
    \includegraphics[width=\columnwidth]{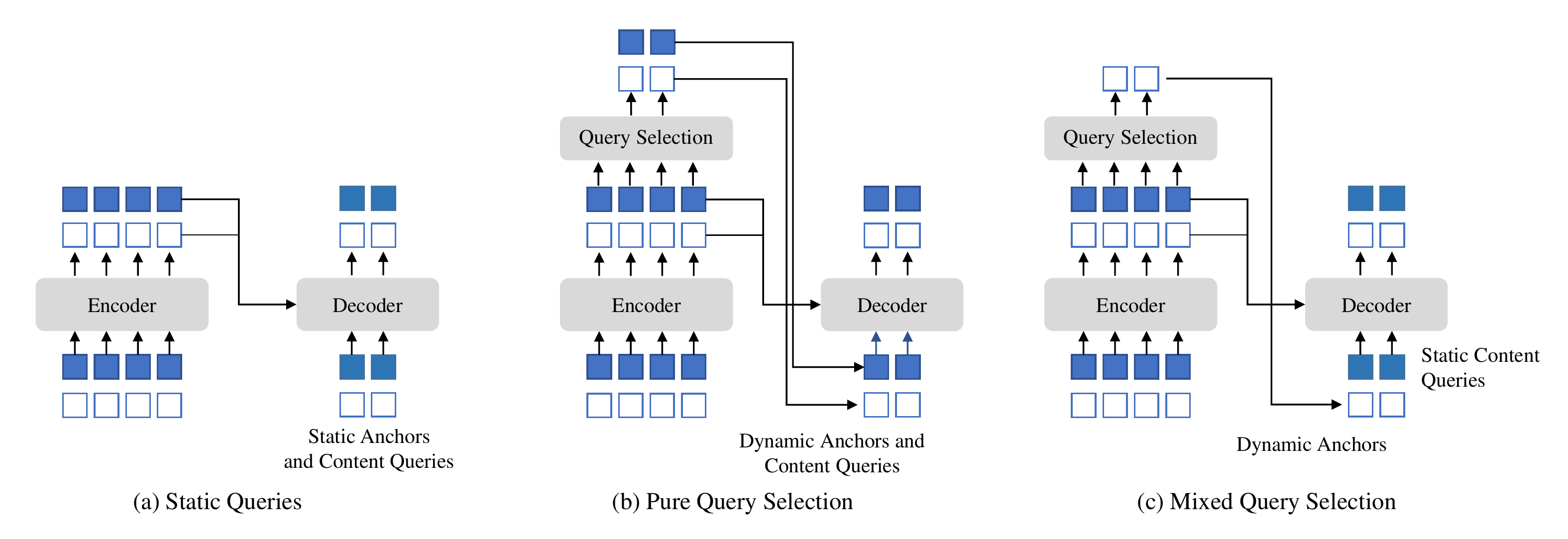}
    \centering
    \caption{Comparison of three different query initialization methods. The term ``static'' means that they will keep the same for different images in inference. A common implementation for these static queries is to make them learnable. }
    \label{fig:query_selection}
\end{figure}

In DETR~\cite{carion2020end} and DN-DETR~\cite{li2022dn}, decoder queries are static embeddings without taking any encoder features from an individual image, as shown in Fig. \ref{fig:query_selection} (a). They learn anchors (in DN-DETR and DAB-DETR) or positional queries (in DETR) from training data directly and set the content queries as all $0$ vectors. 
Deformable DETR~\cite{zhu2020deformable} learns both the positional and content queries, which is another implementation of static query initialization.
To further improve the performance, Deformable DETR~\cite{zhu2020deformable} has a query selection variant (called "two-stage" in \cite{zhu2020deformable}), which select top K encoder features from the last encoder layer as priors to enhance decoder queries. As shown in Fig. \ref{fig:query_selection} (b), both the positional and content queries are generated by a linear transform of the selected features.
In addition, these selected features are fed to an auxiliary detection head to get predicted boxes, which are used to initialize reference boxes. Similarly, Efficient DETR~\cite{yao2021efficient} also selects top K features based on the objectiveness (class) score of each encoder feature. 

The dynamic $4$D anchor box formulation of queries in our model makes it closely related to decoder positional queries, which can be improved by query selection. We follow the above practice and propose a mixed query selection approach. As shown in Fig. \ref{fig:query_selection} (c), we only initialize anchor boxes using the position information associated with the selected top-K features, but leave the content queries static as before.
Note that Deformable DETR~\cite{zhu2020deformable} utilizes the top-K features to enhance not only the positional queries but also the content queries. As the selected features are preliminary content features without further refinement, they could be ambiguous and misleading to the decoder. For example, a selected feature may contain multiple objects or be only part of an object. In contrast, our mixed query selection approach only enhances the positional queries with top-K selected features and keeps the content queries learnable as before. 
It helps the model to use better positional information to pool more comprehensive content features from the encoder.


\vspace{-0.2cm}
\subsection{Look Forward Twice}
\label{sec:look_forward}
\begin{figure}[h]
    \includegraphics[width=\columnwidth]{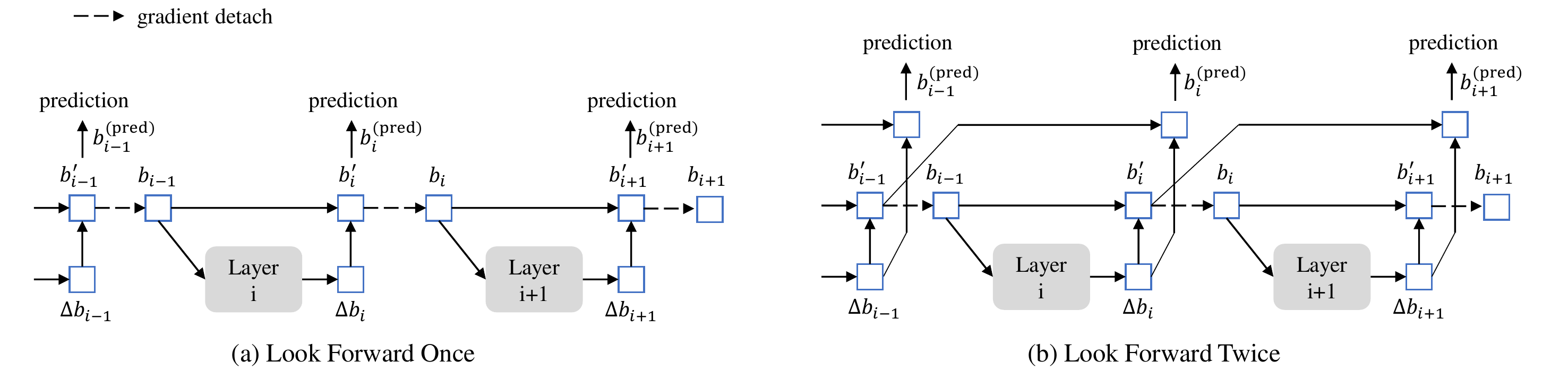}
    \centering
    \caption{Comparison of box update in Deformable DETR and our method.}
    \label{fig:look_forward}
\end{figure}

We propose a new way to box prediction in this section. The iterative box refinement in Deformable DETR~\cite{zhu2020deformable} blocks gradient back propagation to stabilize training. We name the method look forward once since the parameters of layer $i$ are updated based on the auxiliary loss of boxes $b_i$ only, as shown in Fig. \ref{fig:look_forward} (a). 
However, we conjecture that the improved box information from a later layer could be more helpful to correct the box prediction in its adjacent early layer. Hence we propose another way called look forward twice to perform box update, where the parameters of layer-$i$ are influenced by losses of both layer-$i$ and layer-$(i+1)$, as shown in Fig. \ref{fig:look_forward} (b). For each predicted offset $\Delta b_i$, it will be used to update box twice, one for $b_i'$ and another for $b_{i+1}^{(pred)}$, hence we name our method as look forward twice.

The final precision of a predicted box $b_{i}^{(\mathrm{pred})}$ is determined by two factors: the quality of the initial box $b_{i-1}$ and the predicted offset of the box $\Delta b_i$. The look forward once scheme optimizes the latter only, as the gradient information is detached from layer-$i$ to layer-$(i-1)$. In contrast, we improve both the initial box $b_{i-1}$ and the predicted box offset $\Delta b_i$. A simple way to improving the quality is supervising the final box $b_i'$ of layer $i$ with the output of the next layer $\Delta b_{i+1}$. Hence we use the sum of $b_i'$ and $\Delta b_{i+1}$ as the predicted box of layer-$(i+1)$.

More specifically, given an input box $b_{i-1}$ for the $i$-th layer, we obtain the final prediction box  $b_{i}^{(\mathrm{pred})}$ by:

\begin{equation}
\vspace{-0.1cm}
\begin{aligned} 
     \Delta b_{i} &= \mathrm{Layer_{i}}(b_{i-1}), \quad\quad
     &b_{i}' &= \mathrm{Update}(b_{i-1}, \Delta b_{i}), \\
     b_{i} &= \mathrm{Detach}(b_{i}'), \quad\quad
     &b_{i}^{(\mathrm{pred})} &= \mathrm{Update}(b_{i-1}', \Delta b_{i}),
\end{aligned}
\end{equation}

\noindent
where $b_{i}'$ is the undetached version of $b_{i}$. The term $\mathrm{Update}(\cdot,\cdot)$ is a function that refines the box $b_{i-1}$ by the predicted box offset $\Delta b_{i}$. We adopt the same way for box update\footnote{We use normalized forms of boxes in our model, hence each value of a box is a float between $0$ and $1$. Given two boxes, we sum them after inverse sigmoid and then transform the summation by sigmoid. Refer to Deformable DETR~\cite{zhu2020deformable} Sec. A.3 for more details.} as in Deformable DETR~\cite{zhu2020deformable}. 

\section{Experiments}
\subsection{Setup}\label{sec:setup}
\textbf{Dataset and Backbone:}
We conduct evaluation on the COCO 2017 object detection dataset~\cite{lin2015microsoft}, which is split into \texttt{train2017} and \texttt{val2017} (also called \texttt{minival}). We report results with two different backbones: ResNet-50~\cite{he2015deep} pre-trained on ImageNet-1k~\cite{deng2009imagenet} and SwinL~\cite{liu2021swin} pre-trained on ImageNet-22k~\cite{deng2009imagenet}. 
DINO with ResNet-50 is trained on \texttt{train2017} without extra data, while DINO with SwinL is first pre-trained on Object365~\cite{shao2019objects365} and then fine-tuned on \texttt{train2017}.
We report the standard average precision (AP) result on \texttt{val2017} under different IoU thresholds and object scales. We also report the \texttt{test-dev} results for DINO with SwinL.
\\
\textbf{Implementation Details:}
 DINO is composed of a backbone, a Transformer encoder, a Transformer decoder, and multiple prediction heads. 
In appendix \ref{sec:more_imp_details}, we provide more implementation details, including all the hyper-parameters and engineering techniques used in our models for those who want to reproduce our results. We will release the code after the blind review.

\subsection{Main Results}
\begin{table}[h]
    \centering
    \resizebox{\textwidth}{!}{%
    \begin{tabular}{l|c|cccccc|c|c|c}
        \shline
        Model  & Epochs & AP & AP$_{50}$ & AP$_{75}$ & AP$_{S}$ & AP$_{M}$ & AP$_{L}$ & GFLOPS & Params&FPS \\
        \shline
        Faster-RCNN(5scale) \cite{ren2015faster}& $12$ & $37.9$ & $58.8$ & $41.1$ & $22.4$ & $41.1$ & $49.1$ & $207$ & $40$M & $21^{*}$ \\
        DETR(DC5) \cite{carion2020end}& $12$ & $15.5$ & $29.4$ & $14.5$ & $4.3$ & $15.1$ & $26.7$ &$225$ & $41$M  & $20$ \\
        Deformable DETR(4scale)\cite{zhu2020deformable}& $12$ & $41.1$ &  $-$ & $-$ & $-$ & $-$ & &$196$ & $40$M & 24\\
        DAB-DETR(DC5)$^\dag$ \cite{liu2022dab}& $12$ & $38.0$ & $60.3$ & $39.8$ & $19.2$ & $40.9$ & $55.4$ &$256$ & $44$M & $17$ \\
        Dynamic DETR(5scale) \cite{Dai_2021_ICCV}& $12$ & $42.9$ & $61.0$ & $46.3$ & $24.6$ & $44.9$ & $54.4$ &$-$ & $58$M&$-$\\
        Dynamic Head(5scale) \cite{dai2021dynamic}& $12$ & $43.0$ &$60.7$ & $46.8$ & $24.7$ & $46.4$ & $53.9$ &$-$ & $-$&$-$\\
        HTC(5scale) \cite{chen2019hybrid}& $12$ & $42.3$ & $-$ & $-$ & $-$ & $-$ & $-$ &$441$ & $80$M & $5^{*}$ \\
        DN-Deformable-DETR(4scale)$^\dag$ \cite{li2022dn}& $12$ & $43.4$ & $61.9$ & $47.2$ & $24.8$ & $46.8$ & $59.4$ &$265$ & $48$M & $23$\\
        \hline
        DINO-4scale$^\dag$    &12 & $\textbf{49.0}$\fontsize{7.0pt}{\baselineskip}\selectfont${(+5.6)}$& $\textbf{66.6}$ & $\textbf{53.5}$ & $\textbf{32.0}$\fontsize{7.0pt}{\baselineskip}\selectfont${(+7.2)}$ & $\textbf{52.3}$ & $\textbf{63.0}$ &$279$ & $47$M & 24 \\
        DINO-5scale$^\dag$    &12 &$\textbf{49.4}$\fontsize{7.0pt}{\baselineskip}\selectfont${(+6.0)}$&  $\textbf{66.9}$ & $\textbf{53.8}$ & $\textbf{32.3}$\fontsize{7.0pt}{\baselineskip}\selectfont${(+7.5)}$ &$\textbf{52.5}$ & $\textbf{63.9}$ &$860$ & $47$M &$10$\\
        \shline
    \end{tabular}}
    \centering
    \vspace{0.1cm}
    \caption{Results for DINO and other detection models with the ResNet50 backbone on COCO \texttt{val2017} trained with $12$ epochs (the so called $1\times$ setting). For models without multi-scale features, we test their GFLOPS and FPS for their best model ResNet-50-DC5. DINO uses $900$ queries. $^\dag$ indicates models that use $900$ queries or $300$ queries with $3$ patterns which has similar effect with $900$ queries. Other DETR-like models except DETR ($100$ queries) uses $300$ queries.
    $^*$ indicates that they are tested using the mmdetection~\cite{chen2019mmdetection} framework.
    }
    \label{tab:12ep}
\end{table}
\noindent
\textbf{12-epoch setting:}
With our improved anchor box denoising and training losses, the training process can be significantly accelerated. As shown in Table~\ref{tab:12ep}, we compare our method with strong baselines including both convolution-based methods~\cite{ren2015faster,chen2019hybrid,dai2021dynamic} and DETR-like methods~\cite{carion2020end,zhu2020deformable,Dai_2021_ICCV,liu2022dab,li2022dn}. For a fair comparison, we report both GFLOPS and FPS tested on the same A100 NVIDIA GPU for all the models listed in Table~\ref{tab:12ep}. All methods except for DETR and DAB-DETR use multi-scale features. For those without multi-scale features, we report their results with ResNet-DC5 which has a better performance for its use of a dilated larger resolution feature map. 
Since some methods adopt $5$ scales of feature maps and some adopt $4$, we report our results with both $4$ and $5$ scales of feature maps. 

As shown in Table \ref{tab:12ep}, our method yields an improvement of $+5.6$ AP under the same setting using ResNet-50 with $4$-scale feature maps and $+6.0$ AP with $5$-scale feature maps. Our $4$-scale model does not introduce much overhead in computation and the number of parameters. Moreover, our method performs especially well for small objects, gaining $+7.2$ AP with $4$ scales and $+7.5$ AP with $5$ scales. Note that the results of our models with ResNet-50 backbone are higher than those in the first version of our paper due to engineering techniques.
\\
\noindent
\textbf{Comparison with the best models with a ResNet-50 backbone:}
\begin{table}[h]
    \centering
    \resizebox{\textwidth}{!}{%
    \begin{tabular}{l|c|cccccc}
        \shline
        Model  & Epochs & AP & AP$_{50}$ & AP$_{75}$ & AP$_{S}$ & AP$_{M}$ & AP$_{L}$  \\
        \shline
        Faster-RCNN \cite{ren2015faster}& $108$ & $42.0$ & $62.4$ & $44.2$ & $20.5$ & $45.8$ & $61.1$  \\
        DETR(DC5) \cite{zhu2020deformable}   & $500$ & $43.3$ & $63.1$ & $45.9$ & $22.5$ & $47.3$ & $61.1$   \\ 
        Deformable DETR \cite{zhu2020deformable}   & $50$ & $46.2$ & $65.2$ & $50.0$ & $28.8$ & $49.2$ & $61.7$   \\ 
        SMCA-R \cite{gao2021fast}               & $50$ &  $43.7$ &  $63.6$ &  $47.2$ &  $24.2$ &  $47.0$ &  $60.4$  \\
        TSP-RCNN-R \cite{sun2020rethinking}          & $96$ & $45.0$ & $64.5$ & $49.6$ & $29.7$ & $47.7$ & $58.0$  \\
        Dynamic DETR(5scale) \cite{dai2021dynamic}  & $50$ & ${47.2}$ & $65.9$ & $51.1$ & $28.6$ & $49.3$ & $59.1$  \\
        DAB-Deformable-DETR \cite{liu2022dab}     & $50$ & {$46.9$} & $66.0$ & $50.8$ & $30.1$ & $50.4$ & $62.5$  \\
        DN-Deformable-DETR \cite{li2022dn}    &50 &{48.6}& $67.4$ & $52.7$ & $31.0$ & $52.0$ & $63.7$ \\
        \hline
        DINO-4scale    &24 &\textbf{50.4}\fontsize{7.0pt}{\baselineskip}\selectfont{(+1.8)}&$68.3$&$54.8$&$33.3$ &$53.7$&$64.8$ \\
        DINO-5scale    &24 &\textbf{51.3}\fontsize{7.0pt}{\baselineskip}\selectfont{(+2.7)}&$69.1$&$56.0$&$34.5$ &$54.2$&$65.8$  \\
        \hline
        DINO-4scale    &36 &\textbf{50.9}\fontsize{7.0pt}{\baselineskip}\selectfont{(+2.3)}& $69.0$ & $55.3$ & $34.6$ & $54.1$ & $64.6$ \\
        DINO-5scale    &36 &\textbf{51.2}\fontsize{7.0pt}{\baselineskip}\selectfont{(+2.6)}& $69.0$ & $55.8$ & $35.0$ & $54.3$ & $65.3$ \\
        \shline
    \end{tabular}}
    \centering
    \vspace{0.2cm}
    \caption{Results for DINO and other detection models with the ResNet-50 backbone on COCO \texttt{val2017} trained with more epochs ($24$, $36$, or more). 
    }
    
    \label{tab:50ep}
\end{table}
\begin{figure}[h]
    \centering
\includegraphics[width=\columnwidth]{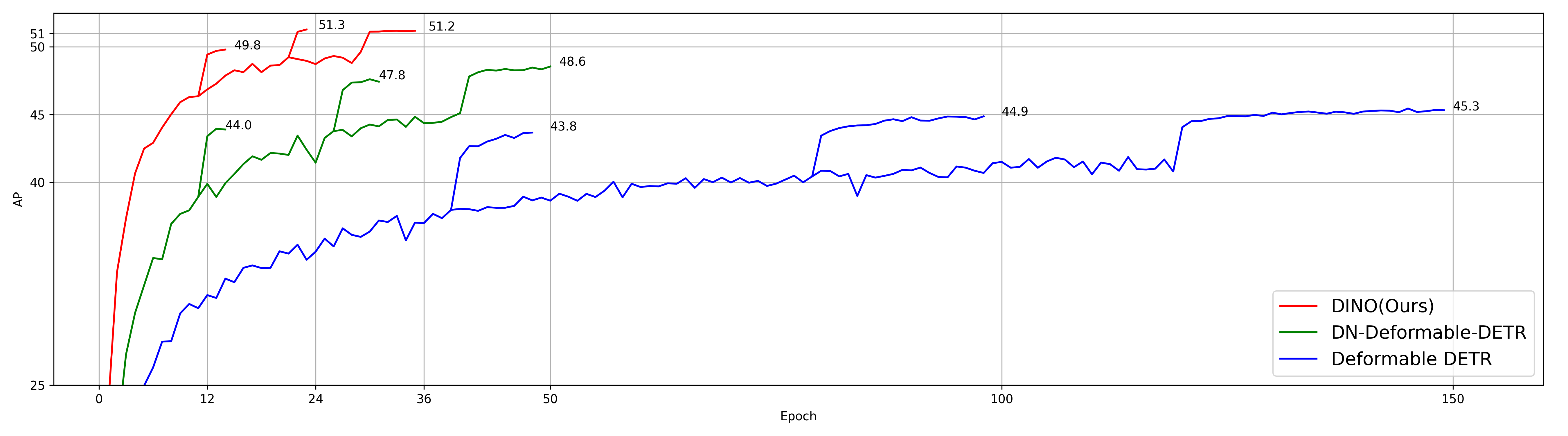}
    \caption{Training convergence curves evaluated on COCO \texttt{val2017} for DINO and two previous state-of-the-art models with ResNet-50 using multi-scale features.}
    \label{fig:convergence_details}
    \vspace{0.1cm}
\end{figure}

To validate the effectiveness of our method in improving both convergence speed and performance, we compare our method with several strong baselines using the same ResNet-50 backbone. Despite the most common $50$-epoch setting, we adopt the $24$ ($2\times$) and $36$ ($3\times$) epoch settings since our method converges faster and yields only a smaller additional gain with $50$-epoch training.
The results in Table~\ref{tab:50ep} show that, using only $24$ epochs, our method achieves an improvement of $+1.8$ AP and $+2.7$ AP with $4$ and $5$ scales, respectively. Moreover, using $36$ epochs in the $3\times$ setting, the improvement increases to $+2.3$ and $+2.6$ AP with $4$ and $5$ scales, respectively. The detailed convergence curve comparison is shown in Fig. \ref{fig:convergence_details}.
\\
\begin{table}[h]
    \centering
        \footnotesize
            \renewcommand{\arraystretch}{1.3}
    \resizebox{1.0\columnwidth}{!}{%
    \begin{tabular}{c|c|c|c|c|c|cc|cc}
        \shline
        \multirow{2}{*}{Method} & \multirow{2}{*}{Params} & \multirow{2}{*}{\makecell{Backbone Pre-training \\ Dataset }} &  \multirow{2}{*}{\makecell{Detection Pre-training \\ Dataset }} & \multirow{2}{*}{\makecell{Use \\Mask}} & \multirow{2}{*}{End-to-end}  & \multicolumn{2}{c|}{{ val2017 (AP)}} & \multicolumn{2}{c}{{test-dev (AP)}} \\
        & & & & & & \scriptsize w/o TTA& \scriptsize w/ TTA &\scriptsize w/o TTA&\scriptsize w/ TTA \\
        \shline
        SwinL~\cite{liu2021swin} & $284$M & IN-22K-14M & O365  & \checkmark & & $57.1$ & $58.0$ & $57.7$ & $58.7$ \\
        DyHead~\cite{dai2021dynamic} & $\geq 284$M & IN-22K-14M & Unknown*  &  &  & $-$ & $58.4$ &$-$ & $60.6$ \\
        Soft Teacher+SwinL~\cite{xu2021end} & $284$M & IN-22K-14M & O365  & \checkmark & &$60.1$ & $60.7$ & $-$ & $61.3$ \\
        GLIP~\cite{li2021grounded} & $\geq 284$M & IN-22K-14M & FourODs~\cite{li2021grounded},GoldG+~\cite{li2021grounded,AishwaryaKamath2021MDETRM} &  & & $-$ & $60.8$ & $-$ & $61.5$\\
        Florence-CoSwin-H\cite{yuan2021florence} & $\geq 637$M & FLD-900M~\cite{yuan2021florence} &  FLD-9M~\cite{yuan2021florence} &  & & $-$ & $62.0$ & $-$ & $62.4$ \\
        SwinV2-G~\cite{liu2021swinv2} & $3.0$B & IN-22K-ext-70M~\cite{liu2021swinv2} & O365 & \checkmark & & $61.9$ & $62.5$ & $-$ & $63.1$\\
        \hline
        DINO-SwinL(Ours) & $\textbf{218}$\textbf{M} & IN-22K-14M & O365 &  & \checkmark& $\textbf{63.1}$ & $\textbf{63.2}$ & $\textbf{63.2}$ & $\textbf{63.3}$ \\
        \shline
    \end{tabular}
    }
    \vspace{0.05cm}
    \caption{Comparison of the best detection models on MS-COCO. Similar to DETR~\cite{carion2020end}, we use the term ``end-to-end'' to indicate if a model is free from hand-crafted components like RPN and NMS. The term ``use mask'' means whether a model is trained with instance segmentation annotations. We use the terms ``IN'' and ``O365'' to denote the ImageNet~\cite{deng2009imagenet} and Objects365~\cite{shao2019objects365} datasets, respectively. Note that ``O365'' is a subset of ``FourODs'' and ``FLD-9M''. * DyHead does not disclose the details of the datasets used for model pre-training.
    }
    \label{tab:sota}
\end{table}
\subsection{Comparison with SOTA Models}
To compare with SOTA results, we use the publicly available SwinL \cite{liu2021swin} backbone pre-trained on ImageNet-22K. We first pre-train DINO on the Objects365 \cite{shao2019objects365} dataset and then fine-tune it on COCO. As shown in Table \ref{tab:sota}, DINO achieves the best results of $63.2$AP and $63.3$AP on COCO \texttt{val2017} and \texttt{test-dev}, which demonstrate its strong scalability to larger model size and data size. Note that all the previous SOTA models in Table \ref{tab:sota} do not use Transformer decoder-based detection heads (HTC++ \cite{chen2019hybrid} and DyHead \cite{dai2021dynamic}). It is the first time that an end-to-end Transformer detector is established as a SOTA model on the leaderboard \cite{paperwithcode}. Compared with the previous SOTA models, we use a much smaller model size ($1/15$ parameters compared with SwinV2-G \cite{liu2021swinv2}), backbone pre-training data size ($1/60$ images compared with Florence), and detection pre-training data size ($1/5$ images compared with Florence), while achieving better results. In addition, our reported performance without test time augmentation (TTA) is a neat result without bells and whistles. These results effectively show the superior detection performance of DINO compared with traditional detectors.

\subsection{Ablation}
\begin{table}[h]
    \centering
    \resizebox{\textwidth}{!}{%
    \begin{tabular}{l|ccc|cccccc}
        \shline
        \#Row  & QS & CDN & LFT & AP & AP$_{50}$ & AP$_{75}$ & AP$_{S}$ & AP$_{M}$ & AP$_{L}$ \\
        \shline
        1. DN-DETR~\cite{li2022dn} & No  & & & $43.4$ & $61.9$ & $47.2$ & $24.8$ & $46.8$ & $59.4$\\
        2. Optimized DN-DETR &  No & & & $44.9$ & $62.8$ & $48.6$ & $26.9$ & $48.2$ & $60.0$ \\
        3. Strong baseline (Row2+pure query selection) & Pure  &  &   & $46.5$ & $64.2$ & $50.4$ & $29.6$ & $49.8$ & $61.0$ \\
        4. Row3+mixed query selection & Mixed  &  &  & $47.0$ & $64.2$ & $51.0$ & $31.1$ & $50.1$ & $61.5$\\
        5. Row4+look forward twice  & Mixed  &  & \checkmark  & $47.4$ & $64.8$ & $51.6$ & $29.9$ & $50.8$ & $61.9$\\
        6. {\modelname} (ours, Row5+contrastive DN)  & Mixed  & \checkmark  &  \checkmark &  \textbf{47.9}&$\textbf{65.3}$ & $\textbf{52.1}$ & $\textbf{31.2}$ & $\textbf{50.9}$ & $\textbf{61.9}$  \\
        \shline
    \end{tabular}}
    \centering
    \vspace{0.3cm}
    \caption{Ablation comparison of the proposed algorithm components. We use the terms ``QS'', ``CDN'', and ``LFT'' to denote ``Query Selection'', ``Contrastive De-Noising Training'', and ``Look Forward Twice'', respectively.
    }
    \label{tab:ablation}
\end{table}
\noindent
\textbf{Effectiveness of New Algorithm Components:} 
To validate the effectiveness of our proposed methods, 
we build a strong baseline with optimized DN-DETR and pure query selection as described in section \ref{sec:preliminary}. 
We include all the pipeline optimization and engineering techniques (see section \ref{sec:setup} and Appendix \ref{sec:more_imp_details}) in the strong baseline. The result of the strong baseline is available in Table \ref{tab:ablation} Row 3.
We also present the result of optimized DN-DETR without pure query selection from Deformable DETR \cite{zhu2020deformable} in Table \ref{tab:ablation} Row 2. 
While our strong baseline outperforms all previous models, 
our three new methods in DINO further improve the performance significantly.
\section{Conclusion}
In this paper, we have presented a strong end-to-end Transformer detector DINO with contrastive denoising training, mixed query selection, and look forward twice, which significantly improves both the training efficiency and the final detection performance. As a result, DINO outperforms all previous ResNet-50-based models on COCO \texttt{val2017} in both the $12$-epoch and the $36$-epoch settings using multi-scale features. Motivated by the improvement, we further explored to train DINO with a stronger backbone on a larger dataset and achieved a new state of the art, $63.3$ AP on COCO 2017 \texttt{test-dev}. This result establishes DETR-like models as a mainstream detection framework, not only for its novel end-to-end detection optimization, but also for its superior performance.

\clearpage
%
%
{\small
\bibliographystyle{plain}
\bibliography{egbib}
}
\newpage
\appendix
\section{Test Time Augmentations (TTA)}
\label{section:tta}
We aim to build an end-to-end detector that is free from hand-crafted components. However, to compare with traditional detection models, we also explore the use of TTA in DETR-like models. We only use it in our large model with the SwinL backbone. Our TTA does not obtain an inspiring gain compared with traditional detectors, but we hope our exploration may provide some insights for future studies.

We adopt multi-scale test and horizontal flip as TTA. However, the way of ensembling different augmentations in our method is different from that in traditional methods which usually output duplicate boxes. In traditional methods, the ensembling is done by first gathering predictions from all augmentations and ranked by a confidence score. Then, duplicate boxes are found and eliminated by NMS or box voting. The reason why predictions from all augmentations are gathered first is that duplicate boxes appear not only among different augmentations but also within one augmentation. This ensembling method decreases the performance for our method since DETR-like methods are not prone to output duplicate boxes since their set-based prediction loss inhibits duplicate predictions and 
ensembling may incorrectly remove true
positive predictions~\cite{carion2020end}. To address this issue, we designed a one-to-one ensembling method. Assume we have $n$ augmentations $Aug_0, Aug_1, ..., Aug_{n-1}$, where $Aug_i$ has predictions $\mathbf{O}^i$ and a pre-defined hyper-parameter weight $w^i$. $\mathbf{O}^i=\left\{(b^i_0,l^i_0,s^i_0),(b^i_1,l^i_1,s^i_1), ...,(b^i_{m-1},l^i_{m-1},s^i_{m-1})\right\}$ where $b^i_j, l^i_j$ and $s^i_j$ denote the $j$-th boundbox, label and score, respectively. We let $Aug_0$ be the main augmentation which is the most reliable one. For each prediction in $\mathbf{O}^0$, we select the prediction with the highest $IOU$ from predictions of each of other augmentations $\mathbf{O}^1,..., \mathbf{O}^{n-1}$ and make sure the $IOU$ is higher than a predefined threshold. Finally, we ensemble the selected boxes through weighted average as follows
\begin{equation}
    b=\frac{1}{\sum I^i}\sum_{i=o}^{n-1} I^iw^is^i_{idx(i)}b^i_{idx(i)}
\end{equation}
where $I^i=1$ when there is at least one box in $\mathbf{O}^i$ with IOU higher than the threshold and $I^i=0$ otherwise. $idx(i)$ denotes the index of the selected box in $\mathbf{O}^i$.

\section{Training Efficiency}
We provide the GPU memory and training time for our base model in Table \ref{tab:gpu_memory}. All results are reported on 8 Nvidia A100 GPUs with ResNet-50~\cite{he2015deep}. The results demonstrate that our models are not only effective but also efficient for training. 
\begin{table}[h]
    \centering
    \begin{tabular}{l|c|c|c|c|c}
	\shline
	Model & Batch Size per GPU & Traning Time & GPU Mem. & Epoch & AP \\
	\shline
	Faster RCNN~\cite{ren2015faster}* & 8 & $\sim 60$min/ep & $13$GB & $108$ & $42.0$ \\
	DETR~\cite{carion2020end} & 8 & $\sim 16$min/ep & $26$GB & $300$  & $41.2$ \\
	Deformable DETR~\cite{zhu2020deformable}$^{\star}$ & 2 & $\sim 55$min/ep & $16$GB & $50$  & $45.4$ \\
	DINO(Ours) & 2 &  $\sim 55$min/ep & $16$GB & $\textbf{12}$  & $\textbf{47.9}$ \\
	\shline
    \end{tabular}
    \vspace{0.3cm}
    \caption{Training efficieny for different models with ResNet-50 backbone. All models are trianed with 8 Nvidia A100 GPUs. All results are reported by us. * The results of Faster RCNN are tested with the mmdetection framework. $^{\star}$ We use the vanilla Deformable DETR without two-stage and bbox refinement during testing.
    }
    \label{tab:gpu_memory}
\end{table}

\section{Additional Analysis on our Model Components}\label{sec:num_layer}
\begin{table}[h]
    \centering
    \resizebox{0.8\textwidth}{!}{%
    \begin{tabular}{l|c|c|c|c|c|c|c|c}
        \shline
        \# Encoder/Decoder&$6/6$&$4/6$&$3/6$&$2/6$&6/4&6/2&$2/4$&$2/2$  \\
        \shline
        AP &$47.4$& $46.2$&$45.8$&$45.4$&$46.0$&$44.4$&$44.1$&$41.2$   \\
        \shline
    \end{tabular}}
    \centering
    \vspace{0.3cm}
    \caption{Ablation on the numbers of encoder layers and decoder layers with the ResNet-50 backbone on COCO \texttt{val2017}. We use the $12$-epoch setting and $100$ DN queries without negative samples here.
    }
    \label{tab:enc_dec_num}
\end{table}
\noindent
\textbf{Analysis on the Number of Encoder and Decoder Layers: }We also investigate the influence of varying numbers of encoder and decoder layers. As shown in Table \ref{tab:enc_dec_num}, decreasing the number of decoder layers hurts the performance more significantly. For example, using the same $6$ encoder layers while decreasing the number of decoder layers from $6$ to $2$ leads to a $3.0$ AP drop. This performance drop is expected as the boxes are dynamically updated and refined through each decoder layer to get the final results. Moreover, we also observe that compared with other DETR-like models like Dynamic DETR \cite{dai2021dynamic} whose performance drops by $13.8$AP ($29.1$ vs $42.9$) when decreasing the number of decoder layers to $2$, the performance drop of DINO is much smaller. This is because our mixed query selection approach feeds the selected boxes from the encoder to enhance the decoder queries. Therefore, the decoder queries are well initialized and not deeply coupled with decoder layer refinement. 
\begin{table}[h]
    \centering
    \resizebox{0.9\textwidth}{!}{%
    \begin{tabular}{l|c|c|c|c|c|c|c}
        \shline
        \# Denoising query&100 CDN&1000 DN&200 DN&100 DN&50 DN&10 DN&No DN  \\
        \shline
        AP &$47.9$&$47.6$&$47.4$&$47.4$& $46.7$&$46.0$&$45.1$   \\
        \shline
    \end{tabular}}
    \centering
    \vspace{0.3cm}
    \caption{Ablation on number of denoising queries with the ResNet-50 backbone on COCO validation. Note that $100$ CND query pairs contains $200$ queries which are $100$ positive and $100$ negative queries.
    }
    \label{tab:dn_num}
\end{table}
\\
\textbf{Analysis on Query Denoising:} We continue to investigate the influence of query denoising by varying the number of denoising queries. We use the optimized dynamic denoising group (detailed in Appendix \ref{sec:ddg}). As shown in Table \ref{tab:dn_num}, when we use less than $100$ denoising queries, increasing the number can lead to a significant performance improvement. However, continuing to increase the DN number after $100$ yields only a small additional or even worse performance improvement. 
We also analysis the effect of the number of encoder and decoder Layers in Appendix \ref{sec:num_layer}.

\section{More Implementation Details}\label{sec:more_imp_details}
\subsection{Dynamic DN groups}\label{sec:ddg}
In DN-DETR, all the GT objects (label+box) in one image are collected as one GT group for denoising. To improve the DN training efficiency,  multiple noised versions of the GT group in an image are used during training. In DN-DETR, the number of groups is set to five or ten according to different model sizes. As DETR-like models adopt mini-batch training, the total number of DN queries for each image in one batch is padded to the largest one in the batch. Considering that the number of objects in one image in COCO dataset ranges from $1$ to $80$, this design is inefficient and results in excessive memory consumption. To address this problem, we propose to fix the number of DN queries and dynamically adjust the number of groups for each image according to its number of objects.

\subsection{Large-Scale Model Pre-trianing}
Objects365~\cite{shao2019objects365} is a large-scale detection data set with over $1.7M$ annotated images for training and $80,000$ annotated images for validation. To use the data more efficiently, We select the first $5,000$ out of $80,000$ validation images as our validation set and add the others to training.
We pre-train DINO on Objects365 for $26$ epochs using $64$ Nvidia A100 GPUs and fine-tune the model on COCO for $18$ epochs using $16$ Nvidia A100 GPUS. Each GPU has a local batch size of $1$ image only. In the fine-tuning stage, we enlarge the image size to $1.5\times$ (i.e., with max size $1200\times 2000$). This adds around $0.5$ AP to the final result. To reduce the GPU memory usage, we leverage checkpointing~\cite{chen2016training} and mixed precision~\cite{micikevicius2018mixed} during training. Moreover, we use $1000$ DN queries for this large model.

\subsection{Other Implementation Details}\label{sec:imp_details}

\subsubsection{Basic hyper-parameters.}
For hyper-parameters, as in DN-DETR, we use a $6$-layer Transformer encoder and a $6$-layer Transformer decoder and $256$ as the hidden feature dimension. We set the initial learning rate (lr) as $1\times 10^{-4}$ and adopt a simple lr scheduler, which drops lr at the $11$-th, $20$-th, and $30$-th epoch by multiplying 0.1 for the $12$, $24$, and $36$ epoch settings with RestNet50, respectively. We use the AdamW~\cite{kingma2014adam,loshchilov2017decoupled} optimizer with weight decay of $1\times 10^{-4}$ and train our model on Nvidia A100 GPUs with batch size $16$. Since DN-DETR \cite{li2022dn} adopts $300$ decoder queries and $3$ patterns~\cite{wang2021anchor}, we use $300\times 3=900$ decoder queries with the same computation cost. Learning schedules of our DINO with SwinL are available in the appendix.

\subsubsection{Loss function.}
We use the L1 loss and GIOU~\cite{rezatofighi2019generalized} loss for box regression and focal loss~\cite{lin2018focal} with $\alpha=0.25,\gamma=2$ for classification. As in DETR~\cite{carion2020end}, we add auxiliary losses after each decoder layer. Similar to Deformable DETR~\cite{zhu2020deformable}, we add extra intermediate losses after the query selection module, with the same components as for each decoder layer. We use the same loss coefficients as in DAB-DETR~\cite{liu2022dab} and DN-DETR~\cite{li2022dn}, that is, $1.0$ for classification loss, $5.0$ for L1 loss, and $2.0$ for GIOU loss.

\subsubsection{Detailed model components.}
We also optimize the detection pipeline used in DAB-DETR~\cite{liu2022dab} and DN-DETR~\cite{li2022dn}. Following DN-Deformable-DETR~\cite{li2022dn}, we use the same multi-scale approach as in Deformable DETR~\cite{zhu2020deformable} and adopt the deformable attention. DN-DETR uses different prediction heads with unshared parameters in different decoder layers. In addition, we introduce dynamic denoising group to increase denoising training efficiency and alleviate memory overhead (see Appendix \ref{sec:ddg}). In this work, we find that using a shared prediction head will add additional performance improvement. This also leads to a reduction of about one million parameters. In addition, we find the conditional queries~\cite{meng2021conditional} used in DAB-DETR does not suit our model and we do not include them in our final model. 

\subsubsection{Training augmentation.}
We use the same random crop and scale augmentation during training following DETR~\cite{carion2020end}. For example, we randomly resize an input image with its shorter side between $480$ and $800$ pixels and its longer side at most $1333$. For DINO with SwinL, we pre-train the model using the default setting, but finetune using $1.5\times$ larger scale (shorter side between $720$ and $1200$ pixels and longer side at most $2000$ pixels) to compare with models on the leaderboard~\cite{paperwithcode}. Without using any other tricks, we achieve the result of $63.1$ on \texttt{val2017} and $63.2$ on \texttt{test-dev} without test time augmentation (TTA) (see Appendix \ref{section:tta}), outperforming  the previous state-of-the-art result $63.1$ achieved by SwinV2~\cite{liu2021swinv2} with a much neater solution. 

\subsubsection{Multi-scale setting.}
For our $4$-scale models, we extract features from stages $2$, $3$, and $4$ of the backbone and add an extra feature by down-sampling the output of the stage $4$. An additional feature map of the backbone stage $1$ is used for our $5$-scale models.
For hyper-parameters, we set $\lambda_1=1.0$ and $\lambda_2=2.0$ and use $100$ CDN pairs which contain $100$ positive queries and $100$ negative queries.
\subsection{Detailed Hyper-parameters}
We list the hyper-parameters for those who want to reproduce our results in Table \ref{tab:hyperparameters}.

\begin{table}[]
    \centering
    \begin{tabular}{l|c}
	\shline
	Item & Value \\
	\shline
	lr & 0.0001 \\ \hline
	lr\_backbone & 1e-05 \\ \hline
	weight\_decay & 0.0001 \\ \hline
	clip\_max\_norm & 0.1 \\ \hline
	pe\_temperature & 20 \\ \hline
	enc\_layers & 6 \\ \hline
	dec\_layers & 6 \\ \hline
	dim\_feedforward & 2048 \\ \hline
	hidden\_dim & 256 \\ \hline
	dropout & 0.0 \\ \hline
	nheads & 8 \\ \hline
	num\_queries & 900 \\ \hline
	enc\_n\_points & 4 \\ \hline
	dec\_n\_points & 4 \\ \hline
	transformer\_activation & ``relu'' \\ \hline
	batch\_norm\_type & ``FrozenBatchNorm2d'' \\ \hline
	set\_cost\_class & 2.0 \\ \hline
	set\_cost\_bbox & 5.0 \\ \hline
	set\_cost\_giou & 2.0 \\ \hline
	cls\_loss\_coef & 1.0 \\ \hline
	bbox\_loss\_coef & 5.0 \\ \hline
	giou\_loss\_coef & 2.0 \\ \hline
	focal\_alpha & 0.25 \\ \hline
	dn\_box\_noise\_scale & 0.4 \\ \hline
	dn\_label\_noise\_ratio & 0.5 \\ \hline
	\shline
    \end{tabular}
    \vspace{0.2cm}
    \caption{Hyper-parameters used in our models.}
    \label{tab:hyperparameters}
\end{table}

\end{document}